\documentclass{article}

\usepackage[T1]{fontenc}
\usepackage[utf8]{inputenc}
\usepackage{lmodern}
\usepackage{arxiv}

\usepackage{amsmath}
\usepackage{amssymb}
\usepackage{booktabs}
\usepackage{graphicx}
\usepackage[most]{tcolorbox}
\usepackage[numbers]{natbib}
\usepackage{hyperref}
\hypersetup{colorlinks=true, linkcolor=black, citecolor=black, urlcolor=blue}

\graphicspath{{Fig/}}

\renewcommand{\shorttitle}{LLMs as theory coders}

\newcounter{boxnum}
\renewcommand{\theboxnum}{\arabic{boxnum}}
\newenvironment{boxtext}[1][]{%
  \par\medskip\refstepcounter{boxnum}%
  \begin{tcolorbox}[breakable, enhanced, colback=black!3, colframe=black!55,
      boxrule=0.6pt, arc=2pt, left=10pt, right=10pt, top=8pt, bottom=8pt,
      coltitle=black, colbacktitle=black!8, fonttitle=\bfseries,
      title={Box~\theboxnum.\ \ #1}]%
}{%
  \end{tcolorbox}\par\medskip%
}

\title{Correct codes for the wrong reasons? validating LLMs as measurement instruments for theoretical constructs}

\author{%
  Manuel Pita\thanks{To whom correspondence should be addressed:
    \href{mailto:manuel.pita@ulusofona.pt}{manuel.pita@ulusofona.pt}.\ \
    ORCID:~\href{https://orcid.org/0000-0003-2180-6823}{0000-0003-2180-6823}.}\\
  Artificial Intelligence, Social Interaction and Complexity Laboratory\\
  CICANT, Universidade Lus\'{o}fona\\
  376, Campo Grande, 1700-097 Lisbon, Portugal\\
  \texttt{manuel.pita@ulusofona.pt}
}

\date{\today}

\begin{document}
\maketitle

\begin{abstract}
When a large language model (LLM) codes a construct in text as a human annotator would, that agreement makes the LLM a reliable coder. Yet reliability leaves construct validity untouched. The instrument may be theory-naive, reaching the code through a correlate that meets none of the demands the construct's theory makes, and no current method tells that apart from genuine measurement. We propose grain calibration as a method that closes the gap. It decomposes a construct into clause-level components, tests each against the text with extractive evidence, and combines the results through an explicit, theory-derived rule. Because the rule is stated rather than lodged in one opaque pass, its structure is evidence about the process rather than the output. It shows which components settled a code, and, when the code is wrong, whether a component was missed or an adjacent construct mistaken for it. Validation shifts from scoring an instrument's outputs against an annotator to showing that the instrument runs on the construct its theory specifies.
\end{abstract}



\section{LLMs as coding instruments}\label{sec:problem}


One of the instruments that turns text into data across the social and behavioural sciences is quietly changing hands.
Measuring a construct in text---the morality in a film review, stance in a reply to a group chat, or aspirations in an interview---traditionally meant training human coders on the inferences the construct requires.
Many are turning to using large language models (LLMs) instead.
They often agree with the human coders, even as closely as expert coders agree with one another.
But there is a consequential paradox in the emerging picture of LLMs as coding instruments: they can agree with humans without actually `measuring' the construct's theoretical demands.
This paper maps how \textit{agreement} came to stand in for \textit{validity} when LLMs are used as coding instruments, and proposes grain calibration, an iterative procedure that tunes the instrument until its codes meet the demands of the construct's theory.

Whether LLMs can `reason' is contested.
A recent synthesis finds it premature to decide whether they hold structured internal representations \cite{pavlick:2023}, and work on capability evaluation warns against inferring such capacities from task performance alone \cite{mitchell:2026}.
We stay deliberately agnostic on the matter.
What follows asks not what an LLM is internally, but what it does as a coding instrument.
If we want to know what an LLM is doing, one way is to study it as we would any instrument whose internal mechanisms are sealed.
This often means designing specific inputs, feeding them to the coding instrument, and analysing what it returns.

But why study an instrument that works? The success record of LLMs as reliable coders is substantial by now.
Handed tweets and a two-sentence prompt, GPT is more self-consistent across coding runs than trained coders are with each other, and outperforms crowd workers by $25\%$~\cite{gilardi2023chatgpt}.
Asked to classify political affiliation, GPT-4 reaches $93\%$ accuracy and beats expert coders, picking up oblique cues a supervised classifier misses, from Bible quotations read as conservative signalling to Down syndrome awareness read as pro-life Republican positioning~\cite{tornberg2024chatgpt}.
On standard coding tasks, such as stance and hate-speech classification, zero-shot LLMs reach fair-to-substantial agreement with human coders~\cite{ziems2024can}.
Similarly, in interpretive qualitative coding, a capable LLM can achieve human-level reliability, with chain-of-thought prompting further refining the task~\cite{dunivin2024scalable}.
For sentiment polarity, GPT shows competitive performance, with correlations with human coders as high as $r = 0.77$, while for coding discrete emotions, it achieves $F1$ scores of around $ 0.72$ to $0.78$~\cite{rathje2024gpt}.
Judging by the test the field actually applies (agreement with trained human coders), the LLM instrument \textit{works}.
%

What can we learn about the workings of the (sealed) LLM coding instrument?
First, the studies above indicate that LLMs rely on variables present in the input text---word patterns matching a topic distribution, sentiment valence/arousal clusters, linguistic cues, and even oblique cues available through transitive word-pattern associations.
We adopt the term \textit{surface feature} \cite{chi:1981} for such a variable, which the LLM literature also calls `form' \cite{bender2020climbing}.
Surface features are not the `deep' properties that must be inferred from relations among features \cite{gentner:1983}.
Second, prompt and input text enter the model together as a single sequence of words, so what the input text is coded to express becomes coupled with the instrument that performs the coding.
Third, what the LLM can measure is largely constrained by the representations and associations it learned before the coding task; prompting can steer that capacity, but does not by itself establish that the construct's theoretical relations are being tested~\cite{min:2022}.
These are all challenges to LLMs as coders of theoretical constructs---posits with deep structure that are not directly observable but must be inferred~\cite{bacharach:1989, maccorquodale:1948}.



\subsection{Where the instrument fails}
\label{sec:failure}


The successes above hide two problems in how the instrument reaches a code, and neither shows up in an agreement score.
The first is a \textit{shortcut}: an LLM can produce a code by reading surface features that merely correlate with a construct, rather than the deeper inferences that define it.
Such a code can agree with human ground truth while measuring something else.
The second is \textit{entanglement}: the prompt is part of the instrument, not a neutral codebook instruction that defines its measurement capacity.
One consequence is that prompt rewording can change the measurement, sometimes to its opposite.

Start with the shortcut.
An LLM coder can match human performance without the codebook, by seizing a surface feature, as ordinary language tasks show directly~\cite{geirhos:2020}.
Min and colleagues replaced the codes in a prompt's worked examples with random ones, and across twelve model configurations performance fell by at most $5\%$.
The examples were not installing the codebook but cueing associations the LLM already held~\cite{min:2022}.
In another study, McCoy and colleagues trained a set of models, including the language model BERT, to judge whether one sentence entails another, and then built a set of test cases in which the usual heuristic---predicting entailment when the sentences share most of their words---was deliberately wrong.
 The models followed the heuristic anyway, and performance dropped below $10\%$, far under the $50\%$ they would reach by random guessing, where expert annotators scored $97\%$~\cite{mccoy:2019}.
Both these studies are based on simple NLP tasks, so they establish the mechanism rather than measure the harder construct-coding case.
But the mechanism is general; a coder can match the codes without inferring from the codebook, and the agreement does not reveal how the LLM arrived at the output codes.

The same shortcut reaches the coding of theoretical constructs, where its cost may be a fabricated finding.
Ashwin and colleagues had four LLMs code 2,407 open-ended interviews with Rohingya refugees and their Bangladeshi hosts against a nineteen-code scheme built by trained sociologists~\cite{ashwin2025bias}.
The best LLM reached a mean $F1 = 0.41$, and its errors were not random.
Instead, the errors aligned with the characteristics of the people interviewed in ten of the nineteen codes.
For low educational ambition, the errors ran $48\%$ more negative for refugees than hosts, and $49\%$ more positive for parents of sons than of daughters, enough to flip the sign of the relationship and report, from the errors alone, a pattern that appears nowhere in the interviews.
A supervised model trained on the same codes showed biased errors in only one of the nineteen and recovered the true relationships.
This rules out task difficulty as the cause.
The instrument was not reading refugee status or a child's sex, but surface features that covary with them.
Its diagnostic value is limited, since Ashwin and colleagues establish the structure of the errors without identifying their mechanism.
They hypothesized that the studied populations are underrepresented in the LLM's training data.

Agreement can result from different coders making the same mistake.
Matz and colleagues had ChatGPT and human raters score speed-date transcripts for romantic attraction, with the outcome known independently---whether the daters exchanged contact details~\cite{matz2026romantic}.
The LLM and human coders somewhat agreed with each other ($r = 0.21$ to $r = 0.35$), certainly more closely than either matched the actual outcome (about $r = 0.12$).
 They all read the same wrong surface feature as a signal, taking negation as rejection, whereas in these transcripts, it weakly predicted attraction instead.
A shared folk theory, not a reading of the attraction, produced the agreement (in a single study, and for prediction rather than coding).
The lesson holds across these cases: for an LLM coding instrument, agreement is reliability, not validity, because it cannot separate a reading of the target construct from a shortcut that mimics it.

The second problem is the prompt---with all other variables fixed, rewording it can reverse the measurement.
Asked a separate yes-or-no question for each moral foundation~\cite{graham:2013a,atari:2023}, GPT-4 recovered about fifteen of every hundred comments human coders had marked as authority (recall $R = 0.15$; GPT-4 Turbo, about five)---the instrument vastly under-coded~\cite{rathje2024gpt}.
Shown all the foundations at once and asked which applied, the same LLM coded far more comments as authority than those that actually expressed the construct (precision $P = 0.13$ to $0.19$), thus reversing to over-coding~\cite{bulla2025moral}.
On one fixed corpus, a different prompt change produces the same reversal. Adding worked examples flipped authority from $96\%$ more often than humans to $38\%$ less~\cite{abdurahman2024perils}.
Authority is codable.
The agreement between the best human annotator and the other coders is $F1 = 0.67$, while the best LLM reaches $F1 = 0.25$~\cite{bulla2025moral}.
An instrument that reverses the direction of error with the wording of the codebook is not measuring the target construct.
The reversal is unlikely to be due to random noise or to task difficulty.
The same prompt variations that flip authority leave care---the foundation LLMs code best---stable across the three studies, with the best human annotator achieving $F1 = 0.87$ for care~\cite{bulla2025moral}.

%

Two claims stand, then: the instrument probably arrives at the output codes through a shortcut that agreement cannot trace, and the prompt that runs is coupled with what it measures to a degree that rewording it changes what the instrument measures in the input text.
What separates the coding an LLM gets right from the coding it gets wrong?

\subsection{The robot in the cave}
\label{sec:robot}

To code a construct is to test the input text against the construct's theoretical demands.
Two texts can share a construct's prototypical vocabulary and express completely different constructs.
What divides such constructs is inferred rather than determined from surface features alone.
Consider the following thought experiment about a boy and his robot, B9, entering a cave and reading an inscription carved into the wall (Box~\ref{box-1}).

\begin{boxtext}[The robot in the cave]
\label{box-1}
`Danger!' B9 says.
The robot and the boy have found an inscription carved into the wall of the cave:

\begin{quote}
\emph{Three walked the deep path.\\
The stone closed behind them where water speaks.\\
Their last breath carries still.\\
Follow their names to the source.}
\end{quote}

B9 concludes that the inscription is a deathly warning.
It has identified the relevant surface features: follow, deep path, stone closed, water speaks, and last breath.
The boy is unconvinced.
He has been taught a distinction from comparative ritual texts.
A warning severs the reader from the dead: turn back, do not follow, do not repeat their fate.
A consecration extends the bond: carry their names, continue their path, keep them present.
The same vocabulary of death serves both.

The boy tests the lines that seem decisive.
Each reads both ways.
`Their last breath carries still' can mean that the lethal air has not yet cleared, or that their spirit endures among the living.
`Follow their names to the source' can send the reader after the dead to the water that drowned them, or after their souls to the source.
Even \textit{follow}, the word B9 folded into the danger, settles nothing on its own.
What settles the interpretation is the relation, taken across the whole inscription.
Nowhere does the inscription turn the reader back or break with the dead, the act a warning requires.
Instead, it holds the dead present and sends the reader on.
The inscription does not warn. It consecrates.
\end{boxtext}

The words in the inscription, the surface features and the associations between them painted one coherent picture.
B9 was not confused about the inscription, and yet it failed.
It treated a classification that requires theoretical distinctions as a task that surface features could settle.
Reading such inscriptions requires task decomposition to separate what surface features alone can determine from what requires theoretical inference.
B9 would first ask whether the passage contains death-related language, a surface-cue matching problem.
It would then ask, separately, how the passage positions the reader toward the dead, a theoretically specified interpretive problem.
A recalibrated B9 keeps the two apart, answers the first, and flags the second as unresolved rather than defaulting to a confident code.

The principle extends beyond the cave.
For some constructs, surface features constitute reliable evidence, sufficient to settle the code, while for others, a correct code needs inferences the words can only point to as evidence.
The thought experiment leaves one empirical question open. \textbf{How much of the text-coding landscape falls on each side of the divide?}

\subsection{A gradient of theoretical structure}
\label{sec:gradient}
%
Consider LLM stance coding---classifying a tweet as for, against, or neutral toward a target.
A tweet that reads `This policy is reckless and will hurt working families' contains negative words aimed at the target: `reckless,' `hurt.'
So the against stance reads off the surface.
But sentiment is a correlate of stance, not the construct.
The same stance can appear in either polarity.
Indeed, a more accurate sentiment classifier is no better at stance~\cite{bestvater2023sentiment}.

Sentiment coded as polarity---positive, negative, neutral---is the simpler case, where evaluative words fix the polarity with no target to misread.
But a more nuanced sentiment construct has a structure that becomes invisible when reduced to polarity.
For instance, compassion and outrage are different, yet both fall at the negative end of sentiment as polarity.
If the instrument is expected to code the specific type of negative sentiment, the coarse-coding inference breaks down.
These distinctions do not follow from surface features alone.

Moral-foundation theory makes the distinction clearer.
The care construct, for instance, requires that (a) a sentient being is presented as suffering; (b) the author treats that suffering as morally relevant; (c) the author takes a stance toward it; and (d) the case is distinguished from fairness, loyalty, or authority.
Without these entities and their relationships, care would often be indistinguishable from negative (harm) sentiment.
The three cases above trace a gradient of theoretical complexity: how much each construct demands beyond what surface features alone can settle.

An LLM coding instrument applies the same capacity to any construct across the gradient; what differs is how much inferential work each construct's coding demands.
Call that capacity \emph{distributional competence}: the inferences a model can draw from the `company words keep' in its training data~\cite{firth1957synopsis}.
Its range is wide, from simple keyword matching to sophisticated, multi-step pattern recognition~\cite{mahowald:2024}.
But its basis is always distributional, and never, on its own, the compositional reasoning a theoretical mechanism requires~\cite{bender2021dangers}.


\subsection{Optimizing the wrong target}
\label{sec:wrong-reasons}


Still within the black-box view of the LLM instrument, a reasonable approach to its failures is to improve the prompt, and the field's toolkit for eliciting agreement offers several methods.
Each reworks what surrounds the coding decision without reaching the inferences the construct requires.
Fine-tuning makes the code the training signal; McCoy and colleagues fine-tuned models on hard cases, and the result was that the LLM followed a narrower heuristic~\cite{mccoy:2019}.
Few-shot examples cue an association the model already holds. Scrambling their labels barely changes the output~\cite{min:2022}.
Chain-of-thought offers grounds without showing the code was reached by applying them~\cite{wei2022chain,lin:2025a}.
A richer codebook delivers the construct's definition but does not demonstrate theoretical engagement, and more instruction can even lower consistency~\cite{halterman2025codebook,herderich:2025}.

Return to the cave.
Each prompting approach hands B9 more resources, and it may read the added context well enough to draw the theoretical boundary the boy draws through inference.
Suppose it agrees with the boy on every inscription.
The boy can still explain on what grounds each text consecrates rather than warns; for him those grounds are the method.
B9 may even offer grounds of its own, but we have no way of knowing it arrived at the code by applying them.

So optimization can raise the agreement, and one of these tools may even produce the ideal prompt that tests the construct: the informed decomposition the cave pointed toward.
From agreement alone, though, that prompt is indistinguishable from one that only found a closer correlate; the grounds of the code stay as hidden as before.
The solution cannot be improving agreement alone.

\subsection{Theory-naive instruments}
\label{sec:theory-naive}

B9's failure is an instance of a general condition we call \textit{theory-naivete}: a coding instrument is theory-naive when its coding process does not pass through the construct's \textit{nomological network}---the system of relations a theory specifies between the construct, its components, and what can be observed~\cite{cronbach1955construct}.
For the cave inscriptions, the network specifies that a `warning' positions the reader away from the dead, while a `consecration' positions the reader toward the dead.
B9's inferential path does not pass through these relations.
It passes through surface features and the patterns among them.
A theory-naive instrument may include a detailed codebook, labelled examples, and chain-of-thought reasoning.
The deficit is not in the instrument's resources but in its inferential path.

When the instrument's codes are correct, they cannot be credited to any theoretical principles because the instrument did not test them.
Failures cannot be diagnosed for the same reason.
Earning validity requires evidence that the coding process engages the construct's components: that each component the theory specifies is tested in its own right, and the results combined into the code by a stated rule.
A construct's components have specific theoretical roles.
For most constructs, \textit{detection} and \textit{distinction} components are essential.
The former defines the observables and relations that trigger the construct's expression; the latter defines observables and relations that resolve ambiguities with possible alternative neighbouring constructs.
Different constructs have distinct components that vary with the nature of the underlying theory.
The cave construct has both detection and distinction: a reference to the dead, which any reading must detect, and the stance toward them, which separates a warning from a consecration.
B9 tested the first and read its verdict straight from it, never testing the second.

\subsection{Meaning is not the missing piece}
\label{sec:towards-calibration}


One response to theory-naivete is that the instrument lacks access to meaning.
Bender and Koller~\cite{bender2020climbing} argue that LLMs trained on linguistic form alone cannot acquire meaning grounded outside the text.
The claim is contested. Probing and causal-intervention studies find that LLM representations encode discrete concepts, compositional structure, and abstract semantic roles, more than a strict form-only reading predicts~\cite{pavlick:2023}.
But this does not reach the gap that matters, between accessing meaning and \textit{testing the relations} a construct specifies.
The coding instruments examined here do not lack access to meaning; they read the codebook, process construct definitions written in natural language, and apply them.

Consider B9's assessment of the inscription. It reconstructs the whole scene, three people drowned in a flooding cave.
B9 still fails because it has no theoretical principle for determining whether the inscription warns the living or consecrates the dead.
Bender and Koller's divide is between form and meaning; the divide relevant to LLMs as coding instruments is between surface correlates and required construct inferences.
An instrument that accesses meaning can still code from patterns that correlate with the construct without testing the relations that define it.
Richer understanding does not close that gap, which marks the outer limit of distributional competence; we argue that decomposing the construct into independently testable units does.

\subsection{The three opacities}
\label{sec:opacities}


The aforementioned decomposition requires breaking the instrument into three parts: (a) the codebook that specifies the construct; (b) an inferential engine that applies it, and (c) a procedure that governs how the engine applies the codebook.
The engine may be a human coder, a dictionary method, or a language model; which inferences a codebook affords is a property of the codebook and the engine together, not of either alone~\cite{davis1993knowledge}.
Each operation addresses a place where the inferential path from input text to code remains hidden from inspection---an \textit{opacity} in the coding process.
We identify three cumulative opacity layers:

\begin{enumerate}

\item \textit{Definitional opacity.}
A coding instrument is definitionally opaque when the construct's theoretical components are not externalized in the codebook the instrument receives.
The instrument receives only the construct's name, and the inferential engine supplies whatever content it already associates with that name.
The researcher cannot determine which theoretical components, if any, that content encodes, because the components were never specified as separable targets.
Closing definitional opacity requires decomposing the construct into the components the theory stipulates for a valid coding and supplying each as an explicit, independently assessable instruction.

\item \textit{Inferential opacity.}
A coding instrument is inferentially opaque when the components have been specified but the instrument does not report which evidence in the text supports each component's score.
The codebook specifies what the instrument must check; the instrument returns a code without demonstrating whether it followed the codebook or coded from surface-feature regularities.
Closing inferential opacity requires that each codebook component be tied to an extractive ground: a specific text span that justifies the score and can be audited independently of the final code.

\item \textit{Compositional opacity.}
An LLM coding instrument is compositionally opaque when the rule combining per-component scores into a final code is not stated.
Closing inferential opacity produces grounded scores for each component.
Those scores must be consolidated into a single construct designation, but if the integration rule is internal to the engine, the researcher cannot determine whether one negative component vetoes the code or whether positive components override it.
Closing compositional opacity requires specifying the integration rule as an explicit function that takes inference scores as inputs to derive the code.

\end{enumerate}

The cumulative structure imposes a fixed sequence.
Inferential opacity cannot be closed until definitional opacity has been closed, because per-component evidence requires components to already exist as separable targets.
Compositional opacity cannot be closed until inferential opacity has been closed, because an integration rule requires grounded scores to decide the final output.
In a definitionally opaque instrument, the construct's name cues content in the LLM that the researcher cannot inspect, leaving the resulting code unexamined.
An instrument that closes all three opacities exposes the construct's components, the text evidence supporting each, and the rule that assembles them into a code, the theory's commitments in auditable form.


Even the approaches that close the most opacities provide no evidence that the coding process engages the construct's components rather than distributional correlates.
We next weigh the field's instruments against this requirement.


\section{Construct validity}\label{sec:state-of-art}


The problem above places a single demand on a valid coding instrument: its codes must arise from the process the construct specifies, not from surface correlates.
This demand predates LLMs.
Messick called it the \emph{substantive aspect} of construct validity~\cite{messick1995validity}.
Set out in 1995 as one of six aspects any construct measure must satisfy, it is a standard this field endorses whenever it invokes construct validity.\footnote{A recent account would ground construct validity for LLMs in Cronbach and Meehl's nomological network rather than in Messick~\cite{freiesleben:2026}. The two are complementary, not rival. The nomological network fixes the construct's inferential machinery, while the substantive aspect requires that an instrument's process engages it; Kane, working within the inferential account, grants that construct inferences presuppose a nomological network in the first place~\cite{kane2013validating}.}
A theory-naive instrument has not earned Messick's substantive aspect.


That standard is no longer overlooked.
A growing body of research treats construct validity as a central question for LLMs~\cite{mitchell:2026,freiesleben:2026,lin:2025a,lin:2025b,bean:2025,barrie2026ai}.
Most of this work, however, concerns a different object---whether a benchmark measures a capability attributed to LLMs, not whether they validly code a construct.
What remains is to operationalize the standard for a coding instrument. 
To our knowledge, this has not yet been done.

%
%
The work reviewed below improves the coding instrument.
Sharper definitions, better diagnostics, and iterative refinement each strengthen it.
Yet all measure success the same way, by agreement or by properties of the output, never by whether the coding process engages the construct's components.
Each is read here against that one standard, the substantive aspect.
%
%
Indeed, a now extensive literature refines how construct definitions and annotation guidelines reach the coding instrument~\cite{barrie2024prompt,kim2025repurposing,mohammadi2025definitions,sainz2024gollie,yin2023instructions,xiao2023supporting,zamai2024slimer,dubourg2024stepbystep}.
These interventions close definitional opacity, giving the model the construct's components rather than the code alone.


Methodological consensus treats concordance with human annotators as the terminal validation step.
Primers, tutorials, and review articles across computational social science prescribe the same sequence: compare LLM codes with human codes, report F1 and Cohen's $\kappa$, and declare the instrument validated~\cite{pangakis2023automated,demszky2023llms,brickman2025assessment,goddard2025ramp}.
The most comprehensive codification, Abdurahman et al.'s primer for evaluating LLMs in social-science research~\cite{abdurahman2025primer}, prescribes six steps: validate against human ground truth, check for demographic bias in misclassification, test prompt robustness, document parameters, handle errors, and repeat coding runs.
Every step operates on agreement or its error structure, never on whether that agreement was reached by engaging the construct's components.

The primer is not naive about the risks, as it checks whether misclassifications covary with speaker demographics, cites cases where models responded to disability vocabulary rather than toxicity, and acknowledges that hybrid approaches may outperform zero-shot prompting.
These are the ingredients of a critique that asks what the instrument measures, not merely whether it agrees.
Yet they are never assembled into that critique.
The primer's worked example validates a hypothetical moral-foundation coding study by reporting accuracy, F1, and $\kappa$. It presents balanced per-class performance as the expected outcome---for a construct on which the same research group documented F1 as low as 0.03 for individual foundations~\cite{abdurahman2024perils}.

\subsection{Still the wrong reasons}


One framework reaches further.
Birkenmaier et al.'s ValiText~\cite{birkenmaier2024valitext} adapts Loevinger's tripartite validity model to computational text analysis, distinguishing substantive evidence (theoretical underpinning), structural evidence (model and output properties), and external evidence (concordance with human annotations).
The framework identifies concordance alone as insufficient and warns that a model with strong structural bias may produce misleading associations with external criteria.
Substantive evidence is deemed mandatory---and diagnosed as the weakest class of validity evidence in the LLM-coding literature.\footnote{The call to hold AI evaluation to this standard is not ours alone. Reviewing cognitive AI benchmarks, Mitchell argues that benchmark accuracy seldom establishes the capability a test purports to measure, and frames the deficit explicitly as a lack of \emph{construct validity}---an independent demand for the same standard from outside measurement theory~\cite{mitchell:2026}.}

While the diagnosis is precise, a solution remains absent.
ValiText's seven substantive-evidence steps require the researcher to document a literature review, justify the operationalization, produce a codebook, report interrater agreement, and justify data collection, method, preprocessing, and level of analysis.
All seven are pre-measurement justifications.
They ask the researcher to argue that the measurement should be valid, not to demonstrate empirically that it is.

The framework's closest approach to empirical process evidence is its feature-inspection step, which asks whether the model's top-weighted features are conceptually aligned with the construct.
But feature inspection operates at the level of individual features, not at the level of the construct's compositional structure.
A moral-foundation classifier can weight features such as the presence of `harm' and `suffering' highly, conceptually aligned features, while failing to distinguish care from fairness.
Feature alignment does not reveal compositional failure.
ValiText identifies the absence of substantive evidence in current practice.
It does not provide a procedure for producing it.

In econometrics, the same gap appears from the other end.
Cristian Espinal Maya does provide a procedure, setting conditions under which an LLM score can serve as a measure of a latent variable and adding a statistical correction for the noise in that score~\cite{espinalmaya:2026}.
But the conditions only certify how the score behaves as a variable, such as that it never references the outcome it will later predict, and the correction only removes the bias that the score's noise introduces into later estimates.
Neither framework asks whether the instrument engages the construct's components.


Three other approaches to the validity problem come closest to producing the evidence Birkenmaier's taxonomy demands: (a) reasoning externalization, (b) iterative refinement, and (c) error diagnosis.
Hou et al.~\cite{hou2024prompt} decompose coding tasks into step-based chain-of-thought prompts, each targeting a single dimension of student annotations on a social-learning platform.
The published prompts reveal what the decomposition operationalizes.
The \textit{Theorizing prompt} checks for opinion markers (`think', `believe', `remember'), while the \textit{Integration prompt} checks for response markers (`This is true', `@').
The decomposition follows the codebook's level definitions rather than the construct's theoretical components.
Whether a student constructs an original argument, Theorizing's definitional criterion, requires comparing the annotation to the source text; the prompt checks for vocabulary instead.
A revealing diagnostic confirms this, since including the source text as additional input \textit{decreased} accuracy.
If the model were engaging the construct's components, context should help; that it hurts indicates a surface feature strategy disrupted by additional text, not a construct-engagement strategy aided by relevant context.


The second approach is epistemically different.
Chausson et al.'s Insight-Inference Loop~\cite{chausson:2026} does not treat concordance as validity.
The paper positions its output as researcher-calibrated labels and invokes Bourdieu, Chamboredon, and Passeron's principle that the objects of sociological investigation are constructed through controlled intervention, not given by data.
That intervention \textit{is} the loop, in which the researcher defines claims as single-clause declarative statements, scores each with a natural-language-inference (NLI) model, calibrates decision thresholds against their own annotations, revises underperforming claims, and repeats.
Its limitation is architectural, not epistemological.
The loop calibrates the instrument against the researcher's holistic judgement of whether a text contains a claim.
But that reference standard shares the instrument's vulnerability.
When the NLI model responds to a component's surface features rather than the component itself, the researcher, seeing the same vocabulary, reaches the same conclusion.
An error both make cannot be calibrated away.

Two structural features compound this.
The entailment score compresses the whole document--claim relationship into a single number that does not decompose into per-component evidence; the revision step pushes claims toward the corpus's language, advising the researcher to `emulate the language used' and to replace abstract claims with more literal ones, not toward what the theory says the construct is.
Suppose a researcher addressed every limitation, deriving claims from the construct's theoretical specification, testing each independently against the component it targets, combining the results through a stated rule from theory, and requiring extractive grounding for each judgement.
The researcher would have replaced every structural element of the coding-reliability architecture.
What remains is the NLI engine.
The engine is not the contribution; the validation architecture is.


Xu et al.~\cite{xu:2026} come nearest the substantive aspect, and from inside the agreement paradigm itself.
They reject the single agreement score and sort annotation errors by where they come from and how far they miss.
From how often humans and the model err together, they then estimate how much of the error the task itself makes unavoidable.
The decomposition characterizes the errors systematically: where they fall, how many, and of what kind.
What it does not test is the route, whether the model reached its label by engaging the construct's components or by reading a correlate of them.
And its estimate of task-inherent error holds only where the human coders themselves read the construct.
Where the coders and the model share a surface cue, the overlap is shared invalidity, which agreement cannot separate from genuine task difficulty.
Drawing that line, separating a correlate the coders happen to share from a construct that is simply hard, is what the substantive aspect requires, and where this diagnostic stops.


What none of these approaches supplies is the procedure that would produce that evidence---decompose the construct into components, test each against the text on its own, and combine the results by a stated rule.


The three opacities reveal the most general types of error.
For example, a coding instrument that relies solely on surface features exhibits compositional opacity, with no stated rule for combining component evidence. 
`Harm' and `suffering' can lead to coding the care moral foundation, regardless of whether a sentient being is presented as harmed, the harm is morally appraised, or the case is distinguished from fairness or loyalty \cite{rathje2024gpt,bulla2025moral,abdurahman2024perils}.
The demographically structured errors documented in qualitative interview coding~\cite{ashwin2025bias} are a consequence of inferential opacity, in which the LLM's generative process leaves no record of which component drove the code, leaving the researcher unable to determine whether errors follow what was said or who said it.


The gap between distributional competence and theoretical structure is not closed by refining what the instrument receives---richer definitions, more examples, externalized reasoning steps, iterative threshold calibration---because the problem is in the instrument's architecture.
%
%
While the opacities locate where theory-naivete operates, they do not prescribe a validation procedure.
We outline a procedure in the next section.


\section{Grain calibration}\label{sec:new-approach}


%
The three opacities specify what the architecture of the LLM coding instrument must make accountable to validation: (a) the components a construct's theory specifies; (b) the inference that tests each one against the text, and (c) the rule that combines those tests into code.
These three architectural elements are derived from the construct's \textit{nomological network}, the relations a theory draws among the construct, its components, and what the text can show.
Calibration makes the instrument traverse the nomological network rather than bypass it, as a theory-naive instrument does.
The decomposition follows the construct-to-component relations; individual clauses test a component-to-text relation; the integration rule encodes how the components compose the construct.
Indeed, the failure literature reviewed above shows that decomposition works: even a post-hoc, bottom-up decomposition of a coding task roughly doubles the variance a model's predictions explain~\cite{matz2026romantic}.

Yet, like agreement with human codes, auditability is not validity either.
An audited path can still be the wrong one.
The components may not be the construct's, the tests may answer an easier question, or the construct's structure may have been replaced by a surface correlate.
Validity is not earned outright but argued from evidence.
\textit{Calibrating} the now-visible structure to the construct (grain) earns the substantive aspect, the missing piece of that argument.

Grain calibration sets the \textit{grain} of the decomposition (how finely the construct is split) so that each piece is a question the LLM can answer reliably, then revises it until the evidence shows the theorized components are the ones doing the work.
This paper outlines the procedure and makes the case that it earns the substantive aspect.
A forthcoming paper formalizes the method and demonstrates it end-to-end on moral-foundations coding.
Suffice to say, the procedure is iterative, with a human in the loop and a stop condition.
It starts from the earlier decomposition, the target construct broken into components, each a condition the theory says an instance must meet.

The method refines one step further, splitting each theoretical component into \textit{clauses}---binary questions about what the text must show, each tied to the span that answers it.
The split does not run to arbitrary depth; it stops where every clause falls within distributional competence, where the question is one the LLM can answer from surface features without the theoretical relation the construct depends on.
Decompose too coarsely, and a clause still hides an inference the engine cannot make, a misfit visible in the instrument's behaviour, not a judgement the analyst makes in advance.
Decompose too finely, and the clauses fragment past what the theory demands, into questions about e.g. language rather than the construct.

Setting the grain is half of the procedure; the other half earns the warrant, the substantive aspect in Messick's terms \cite{messick1995validity}.
Each clause is run on the input text, and its answer is recorded with the verbatim span that justifies it.
A stated rule, explicit and inspectable, rather than lodged in the LLM's single generative pass, combines the clause answers into an output code.
The rule is a model fit to human codes as ground truth with per-clause coefficients (weights).
Note that this is not agreement-maximization.
The inputs are theory-grounded clauses, not surface features, so the weights are read against the construct's nomological network rather than pushed to raise a score.
Through grain calibration, the instrument earns Messick's substantive aspect, not agreement with a criterion, but evidence that the processes the theory posits are the processes the instrument runs.

\subsection{Clauses with grounds}


Not every construct needs grain calibration.
Sentiment as polarity is distributional.
The discriminating signals, valence and arousal, already `live' in the LLM, so an instrument reading only the words codes it directly.
Decomposing it would impose structure the construct lacks.
The MFT care foundation is the contrasting case.
The MFT specifies three components, and the vocabulary of harm satisfies none of them on its own.
A sentient being must be presented as suffering; that suffering must be treated as morally relevant, not an incidental detail; and the text must take an evaluative stance toward it, not a bare report.

Each component becomes a clause---a question put to the text, answered yes or no, with the span that warrants the answer.
Does the text reference suffering or vulnerability?
This clause is \textit{detection}: it specifies what the instrument must catch, and defends against the construct going uncoded where it is present.
Is the welfare of a sentient being at stake, rather than mere negative feeling?
This clause is \textit{distinction}: it specifies what the instrument must refuse, and defends against coding a neighbour---loss, disgust, conflict---as care/harm.
Does the writer take an evaluative stance toward the suffering?
This clause is \textit{appraisal}, the third component, the one an appraisive construct requires; without it, a clinical report of injury would get coded as care.
Detection and distinction guard the two ways an instrument can fail a construct, missing what it covers and admitting what it excludes, the under-representation and irrelevant variance a valid instrument must defend against~\cite{messick1995validity}.

An explicit \textit{integration rule} combines the three answers into the output code.
This is the combination that the single-question prompt cannot make explicit and is subject to calibration, because its rule is buried in the generative pass that produces the output.
The decomposition is not a rival classifier set against the prompt.
It is a diagnostic that makes the prompt's commitments visible, clause by clause, where each can be tested against the construct.


Grain calibration applies to target constructs whose underlying theory defines components through a nomological network, and the possible output codes are defined a priori.
Such constructs are \textit{codable}.
A construct generated from the data rather than specified in advance, such as one built through, for example, grounded theory~\cite{charmaz2014constructing}, has no prior network to decompose, and is uncodable and thus unsuitable for grain calibration.
This is a necessary boundary rather than a limitation.
Grain calibration provides substantive-aspect evidence regarding the construct's theory, not evidence that the theory itself is correct.
A mistaken theory, therefore, yields an instrument that validly operationalizes the wrong construct.

\subsection{Human in the loop}


Return to the cave. B9's verdict was itself a rule, unstated: death-vocabulary present, therefore warning.
Written as weights, that rule puts everything on the reference clause and nothing on the stance, the clause that separates a warning from a consecration.
A rule that weighed the stance would have returned consecration; B9 never tested it.

The weight pattern is the diagnosis, and it holds regardless of how often B9 is right.
An instrument can place all its weight on a vocabulary clause, agree with the boy on most inscriptions, and still never test the stance.
This is the difference between two questions one can put to any coding instrument: whether it agrees with the human codes, and why it reached the code it did.
Agreement answers the first and stops.
The weights answer the second, showing which clause settled the code, and that answer is evidence about the process, the substantive aspect.\footnote{The same move---reading the process rather than scoring the output---appears in abstract reasoning: asked to state the rule behind a correct answer, models reveal that many correct outputs rest on unintended rules, right answers reached by the wrong route~\cite{beger:2025}.}

This is where the human enters the loop.
The researcher reads the fitted weights against what the theory requires.
A weight that departs from it---an essential clause left near zero, or one clause settling the whole code---points to the clause to revise.
The researcher rewrites the clause or the decomposition, re-fits the rule, and reads the weights again.
Calibration is this loop, and the weights are what the analyst turns, iteration by iteration, until every weight is one the theory can warrant.

The form of the rule follows the construct rather than convenience.
A regularized logistic regression serves a present-or-absent construct because its coefficients are readable, its decisions inspectable, and its weights revisable when the theory says a clause should matter more.
A magnitude construct takes a linear rule, and an interaction or threshold enters only when the construct's definition specifies one.
A term added to fit the human codes better, without that warrant, is construct-irrelevant variance at the level of composition, the very thing the rule exists to catch.


Because the clauses are interpretable, so are the fitted rule's errors.
Its confusion matrix, read with the weights, places each misclassification on a clause: an under-weighted detection clause that lets the construct slip, or an under-weighted distinction clause that lets a neighbour in.
The matrix tells the researcher which clause to revise, turning each error from a verdict into an instruction.
Three further readings feed the same revision.
The residual, the gap in agreement between the opaque prompt and the explicit decomposition, measures how much of the prompt's behaviour the components do not yet account for; a large residual signals a correlate the theory omits.
The weight profile, one clause settling the construct while another contributes almost nothing, is a claim about where the construct's weight concentrates, open to dispute.
Collinearity, clauses the theory treats as separate moving together across texts, marks a decomposition cut at the wrong joint.

This raises an objection: has the validity judgement simply moved from the model to the analyst who draws the decomposition?
It has not.
A decomposition is not asserted but tested by the three readings above---cut the construct at the wrong joints and the residual swells, a theory-essential clause is left at zero weight, or separate clauses move together.
The decomposition that survives is not the analyst's preference but a claim the evidence was given every chance to break.

One inference runs the other way, and it is a hypothesis, not a result.
When the disagreement with the human codes is itself structured, concentrated on the clauses the theory makes decisive, the fitted rule can be read as a test of the human coding rather than of the instrument: evidence that the codes rest on a surface correlate the components exclude.
This holds only where the components are independently credited to the construct.
Absent that credit, a structured disagreement indicts the instrument as readily as the coders, so the claim must be argued from the grounding of the components, never from the fit.


\begin{table}[t]
\caption{An LLM coding instrument before and after grain calibration. Each row is a property of the coding process, contrasting a theory-naive instrument with the same instrument after calibration. The codebook is delivered in both columns; what calibration changes is whether the process is decomposed into clauses, combined by a stated rule, and fed back against theory. The last row is the decisive one, reliability alone against evidence for the substantive aspect of validity.\label{tab:current-calibrated}}
\begin{tabular*}{\textwidth}{@{\extracolsep{\fill}}lll@{\extracolsep{\fill}}}
\toprule
 & Theory-naive & Theory-calibrated \\
\midrule
Definitional specification & codebook delivered & codebook delivered \\
Inferential decomposition & opaque single pass & per-component clauses with spans \\
Integration rule & implicit & stated, fitted, read against theory \\
Empirical feedback & agreement ($F_1$) & residual, weights, collinearity \\
Error attribution & opaque & clause-level, under- or over-detection \\
Theory refinement & none & iterative loop \\
Validity evidence & reliability & substantive aspect \\
\bottomrule
\end{tabular*}
\end{table}


Grain calibration is two things at once.
It is a method for operationalizing a theoretical construct---decompose, write the clauses, state the rule---and a procedure for earning its validity, by showing through the weights and the residual that the theorized components are the ones the instrument runs on.
The two are not separable steps.
The same decomposition that builds the instrument is what makes its validity inspectable; construction and warrant are one act seen from two sides.

This is the whole of the contribution, and it is deliberately modest in its raw materials.
There is no new standard of validity here, no new theory of meaning, no claim that language models do or do not understand.
There is a standard the field has held for three decades, and an architecture that, to our knowledge, is the first to produce the evidence that standard demands from an LLM coding instrument (Table~\ref{tab:current-calibrated}).
The provocation of this paper was never that agreement is worthless; it does map to reliability.
The provocation is that the field has been accepting reliability as validity because the procedure that tells them apart had not been specified.
Grain calibration makes the distinction operational---an instrument can now be asked not only whether it agrees, but whether it agrees for the reasons the construct specifies.


%
The grain calibration procedure also defines the boundary of the class of coding problems it does not solve.
Statistical methods that correct a noisy instrument's estimates operate downstream of the code and cannot reveal that the instrument measures the wrong construct; grain calibration operates upstream, on the decomposition that fixes what is measured at all~\cite{egami:2023}.
The remaining problems are concrete rather than foundational, and Box~\ref{box-agenda} outlines two of them.
Each extends the same architecture, from one construct to many, and from a single model to a comparison across models.

The title asked whether an instrument can produce correct codes for the wrong reasons.
It can, but agreement alone will never reveal it.
Grain calibration is what turns that limitation into a validation procedure.
Decompose the construct, calibrate it to what the engine can read, and require the instrument to show that its code rests on the components the theory specifies, not on the words that merely keep them company.


\begin{boxtext}[Research agenda]\label{box-agenda}
\textbf{Cross-construct validation.} Grain calibration predicts that constructs with more theoretical structure leave larger residuals under single-pass coding. Testing this across constructs, from near-distributional ones to densely appraisive ones, would turn the residual into a comparative measure of how much theory a construct demands of its instrument.

\textbf{Multi-model comparisons.} Because the weight profile is read in the theory's own terms, the same decomposition run on different models yields comparable component-process profiles. These profiles offer a construct-level way to compare what models can and cannot reliably code, beyond aggregate agreement.
\end{boxtext}


\section*{Acknowledgements}
The author thanks Daniel Cardoso and Mauricio Martins for helpful discussions.

\section*{Competing interest}
The author declares no competing interests.

\section*{Funding}
This work was supported by the European Union's Horizon Europe research and innovation programme under grant agreement No.\ 101094988 (CRESCINE). Views and opinions expressed are those of the author only and do not necessarily reflect those of the European Union or the granting authority; neither can be held responsible for them.

\section*{Author contributions}
MP conceived the framework, developed the theoretical argument, and wrote the manuscript.

\section*{Data availability}
This article contains no original data. All empirical findings cited are from published sources.


\bibliographystyle{unsrtnat}
\bibliography{references}

@article{chi:1981,
	author = {Chi, Michelene T. H. and Feltovich, Paul J. and Glaser, Robert},
	year = {1981},
	doi = {10.1207/s15516709cog0502_2},
	journal = {Cognitive Science},
	number = {2},
	pages = {121--152},
	title = {Categorization and {{Representation}} of {{Physics Problems}} by {{Experts}} and {{Novices}}},
	volume = {5}}

@article{gentner:1983,
	author = {Gentner, Dedre},
	year = {1983},
	doi = {10.1207/s15516709cog0702_3},
	journal = {Cognitive Science},
	number = {2},
	pages = {155--170},
	title = {Structure-{{Mapping}}: {{A Theoretical Framework}} for {{Analogy}}},
	volume = {7}}

@article{bacharach:1989,
	author = {Bacharach, Samuel B.},
	year = {1989},
	doi = {10.5465/amr.1989.4308374},
	journal = {Academy of Management Review},
	number = {4},
	pages = {496--515},
	title = {Organizational Theories: {{Some Criteria}} for {{Evaluation}}},
	volume = {14}}

@article{maccorquodale:1948,
	author = {MacCorquodale, Kenneth and Meehl, Paul E.},
	year = {1948},
	doi = {10.1037/h0056029},
	journal = {Psychological Review},
	number = {2},
	pages = {95--107},
	title = {On a Distinction between Hypothetical Constructs and Intervening Variables},
	volume = {55}}

@article{chausson:2026,
	author = {Chausson, Sandrine and Fourcade, Marion and Harding, David J. and Ross, Bj{\"o}rn and Renard, Gr{\'e}gory},
	doi = {10.1177/00491241251326819},
	journal = {Sociological Methods \& Research},
	number = {2},
	pages = {568--615},
	title = {The Insight-Inference Loop: Efficient Text Classification via Natural Language Inference and Threshold-Tuning},
	volume = {55},
	year = {2026}}

@inproceedings{egami:2023,
	author = {Egami, Naoki and Hinck, Musashi and Stewart, Brandon M. and Wei, Hanying},
	booktitle = {Advances in Neural Information Processing Systems},
	title = {Using Imperfect Surrogates for Downstream Inference: Design-Based Supervised Learning for Social Science Applications of Large Language Models},
	url = {https://proceedings.neurips.cc/paper_files/paper/2023/hash/d862f7f5445255090de13b825b880d59-Abstract-Conference.html},
	volume = {36},
	year = {2023}}

@article{davis1993knowledge,
	author = {Davis, Randall and Shrobe, Howard and Szolovits, Peter},
	doi = {10.1609/aimag.v14i1.1029},
	journal = {AI Magazine},
	number = {1},
	pages = {17--33},
	title = {What Is a Knowledge Representation?},
	volume = {14},
	year = {1993}}

@article{atari:2023,
	author = {Atari, Mohammad and Haidt, Jonathan and Graham, Jesse and Koleva, Sena and Stevens, Sean T. and Dehghani, Morteza},
	doi = {10.1037/pspp0000470},
	journal = {Journal of Personality and Social Psychology},
	number = {5},
	pages = {1157--1188},
	title = {Morality Beyond the {WEIRD}: How the Nomological Network of Morality Varies Across Cultures},
	volume = {125},
	year = {2023}}

@article{gilardi2023chatgpt,
	author = {Gilardi, Fabrizio and Alizadeh, Meysam and Kubli, Ma{\"e}l},
	doi = {10.1073/pnas.2305016120},
	journal = {Proceedings of the National Academy of Sciences},
	number = {30},
	pages = {e2305016120},
	title = {{ChatGPT} outperforms crowd workers for text-annotation tasks},
	volume = {120},
	year = {2023}}

@article{tornberg2024chatgpt,
	author = {T{\"o}rnberg, Petter},
	doi = {10.1177/08944393241286471},
	journal = {Social Science Computer Review},
	number = {6},
	pages = {1181--1195},
	title = {Large Language Models Outperform Expert Coders and Supervised Classifiers at Annotating Political Social Media Messages},
	volume = {43},
	year = {2025}}

@article{rathje2024gpt,
	author = {Rathje, Steve and Mirea, Dan-Mircea and Sucholutsky, Ilia and Marjieh, Raja and Robertson, Claire E. and Van Bavel, Jay J.},
	doi = {10.1073/pnas.2308950121},
	journal = {Proceedings of the National Academy of Sciences},
	number = {34},
	pages = {e2308950121},
	title = {{GPT} is an effective tool for multilingual psychological text analysis},
	volume = {121},
	year = {2024}}

@article{ziems2024can,
	author = {Ziems, Caleb and Held, William and Shaikh, Omar and Chen, Jiaao and Zhang, Zhehao and Yang, Diyi},
	doi = {10.1162/coli_a_00502},
	journal = {Computational Linguistics},
	number = {1},
	pages = {237--291},
	title = {Can large language models transform computational social science?},
	volume = {50},
	year = {2024}}

@article{barrie2024prompt,
	author = {Barrie, Christopher and Palaiologou, Panagiota and T{\"o}rnberg, Petter},
	journal = {arXiv preprint arXiv:2407.02039},
	title = {Prompt stability scoring for text annotation with large language models},
	year = {2024}}

@inproceedings{bean:2025,
	author = {Bean, Andrew M. and Kearns, Ryan Othniel and Romanou, Angelika and others},
	title = {Measuring What Matters: Construct Validity in Large Language Model Benchmarks},
	booktitle = {Advances in Neural Information Processing Systems 38 (NeurIPS 2025), Datasets and Benchmarks Track},
	year = {2025},
	eprint = {2511.04703},
	archiveprefix = {arXiv}}

@article{lin:2025a,
	author = {Lin, Zhicheng},
	title = {A Validity-Guided Workflow for Robust Large Language Model Research in Psychology},
	journal = {arXiv preprint arXiv:2507.04491},
	year = {2025},
	doi = {10.31234/osf.io/xw98v}}

@article{lin:2025b,
	author = {Lin, Zhicheng},
	title = {From Prompts to Constructs: A Dual-Validity Framework for {LLM} Research in Psychology},
	journal = {arXiv preprint arXiv:2506.16697},
	year = {2025},
	doi = {10.48550/arXiv.2506.16697}}

@book{charmaz2014constructing,
	address = {London},
	author = {Charmaz, Kathy},
	edition = {2nd},
	isbn = {978-0857029140},
	publisher = {Sage},
	series = {Introducing Qualitative Methods},
	title = {Constructing Grounded Theory},
	year = {2014}}

@article{cronbach1955construct,
	author = {Cronbach, Lee J. and Meehl, Paul E.},
	doi = {10.1037/h0040957},
	journal = {Psychological Bulletin},
	number = {4},
	pages = {281--302},
	title = {Construct validity in psychological tests},
	volume = {52},
	year = {1955}}

@article{messick1995validity,
	author = {Messick, Samuel},
	doi = {10.1037/0003-066X.50.9.741},
	journal = {American Psychologist},
	number = {9},
	pages = {741--749},
	title = {Validity of psychological assessment: Validation of inferences from persons' responses and performances as scientific inquiry into score meaning},
	volume = {50},
	year = {1995}}

@article{wei2022chain,
	author = {Wei, Jason and Wang, Xuezhi and Schuurmans, Dale and Bosma, Maarten and Ichter, Brian and Xia, Fei and Chi, Ed and Le, Quoc V. and Zhou, Denny},
	journal = {Advances in Neural Information Processing Systems},
	pages = {24824--24837},
	title = {Chain-of-thought prompting elicits reasoning in large language models},
	volume = {35},
	year = {2022}}

@inproceedings{min:2022,
	address = {Abu Dhabi, United Arab Emirates},
	author = {Min, Sewon and Lyu, Xinxi and Holtzman, Ari and Artetxe, Mikel and Lewis, Mike and Hajishirzi, Hannaneh and Zettlemoyer, Luke},
	booktitle = {Proceedings of the 2022 Conference on Empirical Methods in Natural Language Processing},
	doi = {10.18653/v1/2022.emnlp-main.759},
	pages = {11048--11064},
	publisher = {Association for Computational Linguistics},
	title = {Rethinking the {Role} of {Demonstrations}: {What} Makes In-Context Learning Work?},
	year = {2022}}

@article{pangakis2023automated,
	author = {Pangakis, Nicholas and Wolken, Samuel and Fasching, Neil},
	journal = {arXiv preprint arXiv:2306.00176},
	title = {Automated annotation with generative {AI} requires validation},
	year = {2023}}

@inproceedings{xiao2023supporting,
	author = {Xiao, Ziang and Yuan, Xingdi and Liao, Q. Vera and Abdelghani, Rania and Oudeyer, Pierre-Yves},
	booktitle = {Companion Proceedings of the 28th International Conference on Intelligent User Interfaces ({IUI} '23)},
	doi = {10.1145/3581754.3584136},
	pages = {75--78},
	title = {Supporting qualitative analysis with large language models: Combining codebook with {GPT-3} for deductive coding},
	year = {2023}}

@inproceedings{hou2024prompt,
	author = {Hou, Chenyu and Zhu, Gaoxia and Zheng, Lishan and Huang, Xiaoshan and Zhong, Tianlong and Li, Hanxiang and Du, Han and Ker, Chin Lee},
	booktitle = {Proceedings of the 14th Learning Analytics and Knowledge Conference},
	pages = {518--528},
	title = {Prompt-based and fine-tuned {GPT} models for context-dependent and-independent deductive coding in social annotation},
	year = {2024}}

@article{dunivin2024scalable,
	author = {Dunivin, Zackary Okun},
	journal = {arXiv preprint arXiv:2401.15170},
	title = {Scalable qualitative coding with {LLMs}: Chain-of-thought reasoning matches human performance in some hermeneutic tasks},
	year = {2024}}

@article{dubourg2024stepbystep,
	author = {Dubourg, Edgar and Thouzeau, Valentin and Baumard, Nicolas},
	journal = {Frontiers in Artificial Intelligence},
	pages = {1365508},
	title = {A step-by-step method for cultural annotation by {LLMs}},
	volume = {7},
	year = {2024}}

@article{halterman2025codebook,
	author = {Halterman, Andrew and Keith, Katherine A.},
	doi = {10.1017/pan.2025.10017},
	journal = {Political Analysis},
	title = {Codebook {LLMs}: Evaluating {LLMs} as Measurement Tools for Political Science Concepts},
	year = {2025}}

@inproceedings{kim2025repurposing,
	author = {Kim, Kon Woo and Islamaj, Rezarta and Kim, Jin-Dong and Boudin, Florian and Aizawa, Akiko},
	booktitle = {International Conference on Applications of Natural Language to Information Systems ({NLDB} 2025)},
	pages = {140--151},
	publisher = {Springer},
	series = {Lecture Notes in Computer Science},
	title = {Repurposing Annotation Guidelines to Instruct {LLM} Annotators: A Case Study},
	year = {2025}}

@inproceedings{mohammadi2025definitions,
	author = {Mohammadi, Seyedali and Vedula, Bhaskara Hanuma and Lamba, Hemank and Raff, Edward and Kumaraguru, Ponnurangam and Ferraro, Francis and Gaur, Manas},
	booktitle = {Proceedings of the 2025 Conference on Empirical Methods in Natural Language Processing ({EMNLP})},
	doi = {10.18653/v1/2025.emnlp-main.1648},
	pages = {32380--32393},
	title = {Do {LLMs} Adhere to Label Definitions? {Examining} Their Receptivity to External Label Definitions},
	year = {2025}}

@inproceedings{sainz2024gollie,
	author = {Sainz, Oscar and Garc{\'\i}a-Ferrero, Iker and Agerri, Rodrigo and Lopez de Lacalle, Oier and Rigau, German and Agirre, Eneko},
	booktitle = {Proceedings of the Twelfth International Conference on Learning Representations ({ICLR} 2024)},
	title = {{GoLLIE}: Annotation Guidelines Improve Zero-Shot Information-Extraction},
	year = {2024}}

@inproceedings{yin2023instructions,
	author = {Yin, Fan and Vig, Jesse and Laban, Philippe and Joty, Shafiq and Xiong, Caiming and Wu, Chien-Sheng Jason},
	booktitle = {Proceedings of the 61st Annual Meeting of the Association for Computational Linguistics ({ACL} 2023)},
	title = {Did You Read the Instructions? {Rethinking} the Effectiveness of Task Definitions in Instruction Learning},
	year = {2023}}

@article{zamai2024slimer,
	author = {Zamai, Andrew and Zugarini, Andrea and Rigutini, Leonardo and Ernandes, Marco and Maggini, Marco},
	journal = {arXiv preprint arXiv:2407.01272},
	title = {Show Less, Instruct More: Enriching Prompts with Definitions and Guidelines for Zero-Shot {NER}},
	year = {2024}}

@article{abdurahman2024perils,
	author = {Abdurahman, Suhaib and Atari, Mohammad and Karimi-Malekabadi, Farzan and Xue, Mona J. and Trager, Jackson and Park, Peter S. and Golazizian, Preni and Omrani, Ali and Dehghani, Morteza},
	doi = {10.1093/pnasnexus/pgae245},
	journal = {PNAS Nexus},
	number = {7},
	pages = {pgae245},
	title = {Perils and opportunities in using large language models in psychological research},
	volume = {3},
	year = {2024}}

@article{birkenmaier2024valitext,
	archiveprefix = {arXiv},
	author = {Birkenmaier, Lukas and Wagner, Claudia and Lechner, Clemens},
	eprint = {2307.02863},
	journal = {arXiv preprint},
	title = {{ValiText}: A Unified Validation Framework for Computational Text-Based Measures of Social Constructs},
	url = {https://arxiv.org/abs/2307.02863},
	year = {2023}}

@article{kane2013validating,
	author = {Kane, Michael T.},
	doi = {10.1111/jedm.12000},
	journal = {Journal of Educational Measurement},
	number = {1},
	pages = {1--73},
	title = {Validating the Interpretations and Uses of Test Scores},
	volume = {50},
	year = {2013}}

@article{ashwin2025bias,
	author = {Ashwin, Julian and Chhabra, Aditya and Rao, Vijayendra},
	doi = {10.1177/00491241251338246},
	issn = {0049-1241},
	journal = {Sociological Methods \& Research},
	publisher = {SAGE Publications},
	title = {Using Large Language Models for Qualitative Analysis can Introduce Serious Bias},
	year = {2025}}

@inproceedings{bender2020climbing,
	author = {Bender, Emily M. and Koller, Alexander},
	booktitle = {Proceedings of the 58th Annual Meeting of the Association for Computational Linguistics ({ACL})},
	doi = {10.18653/v1/2020.acl-main.463},
	pages = {5185--5198},
	title = {Climbing towards {NLU}: On Meaning, Form, and Understanding in the Age of Data},
	year = {2020}}

@inproceedings{bender2021dangers,
	author = {Bender, Emily M. and Gebru, Timnit and McMillan-Major, Angelina and Shmitchell, Shmargaret},
	booktitle = {Proceedings of the 2021 {ACM} Conference on Fairness, Accountability, and Transparency ({FAccT})},
	doi = {10.1145/3442188.3445922},
	pages = {610--623},
	title = {On the Dangers of Stochastic Parrots: Can Language Models Be Too Big?},
	year = {2021}}

@article{bulla2025moral,
	author = {Bulla, Luana and De Giorgis, Stefano and Mongiov{\`\i}, Misael and Gangemi, Aldo},
	doi = {10.1016/j.chbr.2025.100609},
	journal = {Computers in Human Behavior Reports},
	pages = {100609},
	title = {Large Language Models meet moral values: A comprehensive assessment of moral abilities},
	volume = {17},
	year = {2025}}

@article{goddard2025ramp,
	author = {Goddard, Alex and Gillespie, Alex},
	doi = {10.1037/met0000787},
	journal = {Psychological Methods},
	title = {The Repeated Adjustment of Measurement Protocols ({RAMP}) Method for Developing High-Validity Text Classifiers},
	year = {2025}}

@article{bestvater2023sentiment,
	author = {Bestvater, Samuel E. and Monroe, Burt L.},
	doi = {10.1017/pan.2022.10},
	journal = {Political Analysis},
	number = {2},
	pages = {235--256},
	publisher = {Cambridge University Press},
	title = {Sentiment is Not Stance: Target-Aware Opinion Classification for Political Text Analysis},
	volume = {31},
	year = {2023}}

@incollection{graham:2013a,
	author = {Graham, Jesse and Haidt, Jonathan and Koleva, Sena and Motyl, Matt and Iyer, Ravi and Wojcik, Sean P. and Ditto, Peter H.},
	booktitle = {Advances in {Experimental Social Psychology}},
	doi = {10.1016/B978-0-12-407236-7.00002-4},
	editor = {Devine, Patricia and Plant, Ashby},
	pages = {55--130},
	publisher = {Academic Press},
	title = {Moral Foundations Theory: {The} Pragmatic Validity of Moral Pluralism},
	volume = {47},
	year = {2013}}

@article{mahowald:2024,
	author = {Mahowald, Kyle and Ivanova, Anna A. and Blank, Idan A. and Kanwisher, Nancy and Tenenbaum, Joshua B. and Fedorenko, Evelina},
	doi = {10.1016/j.tics.2024.01.011},
	issn = {1364-6613},
	journal = {Trends in Cognitive Sciences},
	number = {6},
	pages = {517--540},
	publisher = {Elsevier},
	title = {Dissociating Language and Thought in Large Language Models},
	url = {https://www.sciencedirect.com/science/article/pii/S1364661324000275},
	volume = {28},
	year = {2024}}

@article{pavlick:2023,
	author = {Pavlick, Ellie},
	doi = {10.1098/rsta.2022.0041},
	issn = {1364-503X},
	journal = {Philosophical Transactions of the Royal Society A},
	number = {2251},
	pages = {20220041},
	publisher = {The Royal Society},
	title = {Symbols and Grounding in Large Language Models},
	url = {https://royalsocietypublishing.org/doi/10.1098/rsta.2022.0041},
	volume = {381},
	year = {2023}}

@article{demszky2023llms,
	author = {Demszky, Dorottya and Yang, Diyi and Yeager, David S. and Bryan, Christopher J. and Clapper, Margarett and Chandhok, Susannah and Eichstaedt, Johannes C. and Hecht, Cameron and Jamieson, Jeremy and Johnson, Molly and Jones, Michaela and Krettek-Cobb, Desmond and Lai, Leslie and JonesMitchell, Nirel and Ong, Desmond C. and Dweck, Carol S. and Gross, James J. and Pennebaker, James W.},
	doi = {10.1038/s44159-023-00241-5},
	journal = {Nature Reviews Psychology},
	number = {11},
	pages = {688--701},
	title = {Using large language models in psychology},
	volume = {2},
	year = {2023}}

@article{brickman2025assessment,
	author = {Brickman, Jocelyn and Gupta, Mehak and Oltmanns, Joshua R.},
	doi = {10.1177/25152459251343582},
	journal = {Advances in Methods and Practices in Psychological Science},
	title = {Large Language Models for Psychological Assessment: A Comprehensive Overview},
	year = {2025}}

@article{matz2026romantic,
	author = {Matz, Sandra C. and Peters, Heinrich and Cerf, Moran and Grunenberg, Eric and Eastwick, Paul W. and Back, Mitja D. and Finkel, Eli J.},
	doi = {10.1038/s41598-026-52308-x},
	journal = {Scientific Reports},
	title = {Large language models can detect verbal indicators of romantic attraction},
	year = {2026}}

@article{abdurahman2025primer,
	author = {Abdurahman, Suhaib and Salkhordeh Ziabari, Alireza and Moore, Alexander K. and Bartels, Daniel M. and Dehghani, Morteza},
	doi = {10.1177/25152459251325174},
	journal = {Advances in Methods and Practices in Psychological Science},
	title = {A Primer for Evaluating Large Language Models in Social-Science Research},
	year = {2025}}

@inproceedings{mccoy:2019,
	author = {McCoy, R. Thomas and Pavlick, Ellie and Linzen, Tal},
	booktitle = {Proceedings of the 57th Annual Meeting of the Association for Computational Linguistics},
	doi = {10.18653/v1/P19-1334},
	pages = {3428--3448},
	title = {Right for the Wrong Reasons: Diagnosing Syntactic Heuristics in Natural Language Inference},
	year = {2019}}

@article{geirhos:2020,
	author = {Geirhos, Robert and Jacobsen, J{\"o}rn-Henrik and Michaelis, Claudio and Zemel, Richard and Brendel, Wieland and Bethge, Matthias and Wichmann, Felix A.},
	doi = {10.1038/s42256-020-00257-z},
	journal = {Nature Machine Intelligence},
	number = {11},
	pages = {665--673},
	title = {Shortcut Learning in Deep Neural Networks},
	volume = {2},
	year = {2020}}

@article{beger:2025,
	author = {Beger, Claas and Yi, Ryan and Fu, Shuhao and Denton, Kaleda and Moskvichev, Arseny and Tsai, Sarah and Rajamanickam, Sivasankaran and Mitchell, Melanie},
	doi = {10.48550/arXiv.2510.02125},
	journal = {arXiv preprint arXiv:2510.02125},
	title = {Do {AI} Models Perform Human-Like Abstract Reasoning Across Modalities?},
	year = {2025}}

@article{mitchell:2026,
	author = {Mitchell, Melanie},
	doi = {10.1002/aaai.70061},
	journal = {AI Magazine},
	number = {2},
	pages = {e70061},
	title = {Six Principles for Evaluating Cognitive Capabilities in {AI} Models},
	volume = {47},
	year = {2026}}

@incollection{firth1957synopsis,
	address = {Oxford},
	author = {Firth, John Rupert},
	booktitle = {Studies in Linguistic Analysis},
	pages = {1--32},
	publisher = {Blackwell},
	title = {A Synopsis of Linguistic Theory, 1930--1955},
	year = {1957}}

@article{herderich:2025,
	author = {Herderich, Alina and Lasser, Jana and Galesic, Mirta and Aroyehun, Segun Taofeek and Garcia, David and Garland, Joshua},
	doi = {10.31234/osf.io/tzc9p},
	publisher = {PsyArXiv},
	journal = {PsyArXiv},
	title = {Measuring Complex Constructs in Large-Scale Text with Computational Social Mixed Methods},
	url = {https://doi.org/10.31234/osf.io/tzc9p},
	year = {2025}}

@article{freiesleben:2026,
  author = {Freiesleben, Timo},
  title = {Establishing Construct Validity in {LLM} Capability Benchmarks Requires Nomological Networks},
  year = {2026},
  eprint = {2603.15121},
  journal = {arXiv preprint arXiv:2603.15121},
  doi = {10.48550/arXiv.2603.15121},
  url = {https://arxiv.org/abs/2603.15121}
}

@article{barrie2026ai,
  author = {Barrie, Christopher and Argyle, Lisa P. and Bisbee, James and Heseltine, Michael and Lucas, Christopher and Mellon, Jon and Palmer, Alexis and Roberts, Margaret and Spirling, Arthur},
  title = {{AI} and Research Methods},
  journal = {APSA Preprints (Cambridge Open Engage)},
  year = {2026},
  publisher = {APSA Preprints (Cambridge Open Engage)},
  doi = {10.33774/apsa-2026-h59kk},
  url = {https://doi.org/10.33774/apsa-2026-h59kk}
}

@article{espinalmaya:2026,
  author = {Espinal Maya, Cristian},
  title = {Measuring What Cannot Be Surveyed: {LLMs} as Instruments for Latent Cognitive Variables in Labor Economics},
  year = {2026},
  eprint = {2604.02403},
  journal = {arXiv preprint arXiv:2604.02403},
  doi = {10.48550/arXiv.2604.02403},
  url = {https://arxiv.org/abs/2604.02403}
}

@inproceedings{xu:2026,
  author = {Xu, Zhen and Khatri, Vedant and Dai, Yijun and Liu, Xiner and Li, Siyan and Zhang, Xuanming and Yu, Renzhe},
  title = {Enhancing {LLM}-Based Data Annotation with Error Decomposition},
  year = {2026},
  booktitle = {Proceedings of the International Conference on Learning Analytics and Knowledge (LAK '26)},
  publisher = {ACM},
  eprint = {2601.11920},
  doi = {10.1145/3785022.3785070},
  url = {https://arxiv.org/abs/2601.11920}
}

\end{document}